\documentclass{article}

\usepackage{times}
\usepackage{graphicx} 
\usepackage{subfigure}

\usepackage{natbib}

\usepackage{algorithm}
\usepackage{algorithmic}

\usepackage{hyperref}

\usepackage[accepted]{icml2017}

\usepackage{amsmath}
\usepackage{amsfonts}
\usepackage{booktabs}
\usepackage{pifont}
\usepackage{color}

\definecolor{mygreen}{rgb}{0.4,0.85,0.4}

\newcommand{\pro}{$\alpha$-DiMT}

\def\objML{\mathcal{L}}
\def\objRAML{\mathcal{L}_{(\tau)}}
\def\objRL{\mathcal{L}^{\ast}}
\def\objEnRL{\mathcal{L}_{(\tau)}^{\ast}}
\def\objPro{\mathcal{L}_{(\alpha,\tau)}}
\def\glog{\log_{(\alpha)}}

\def\disPro{p_{\theta}^{(\alpha, \tau)}}
\def\disProTilde{\widetilde{p}_{\theta}^{(\alpha, \tau)}}
\def\ptheta{p_{\theta}}
\def\qtau{q_{(\tau)}}

\icmltitlerunning{Neural Sequence Model Training via $\alpha$-divergence Minimization}

\begin{document}

\twocolumn[
\icmltitle{Neural Sequence Model Training \\via $\alpha$-divergence Minimization}



\icmlsetsymbol{equal}{*}

\begin{icmlauthorlist}
\icmlauthor{Sotetsu Koyamada}{rtc,aist,ku}
\icmlauthor{Yuta Kikuchi}{pfn}
\icmlauthor{Atsunori Kanemura}{aist}
\icmlauthor{Shin-ichi Maeda}{pfn}
\icmlauthor{Shin Ishii}{ku,atr}
\end{icmlauthorlist}

\icmlaffiliation{rtc}{Recruit Technologies Co., Ltd., Tokyo, Japan}
\icmlaffiliation{pfn}{Preferred Networks, Inc., Tokyo, Japan}
\icmlaffiliation{aist}{National Institute of Advanced Industrial Science and Technology (AIST), Tokyo, Japan}
\icmlaffiliation{ku}{Graduate School of Informatics, Kyoto University, Kyoto, Japan}
\icmlaffiliation{atr}{ATR Cognitive Mechanisms Laboratories, Kyoto, Japan}

\icmlcorrespondingauthor{Sotetsu Koyamada}{sotetsu.koyamada@gmail.com}

\icmlkeywords{sequence prediction, neural networks, alpha-divergence, reinforcement learning, maximum likelihood, machine translation}

\vskip 0.3in
]



\printAffiliationsAndNotice{}  

\begin{abstract}
We propose a new neural sequence model training method in which the objective function is defined by $\alpha$-divergence.
We demonstrate that the objective function generalizes the maximum-likelihood (ML)-based and reinforcement learning (RL)-based objective functions as special cases (i.e., ML corresponds to $\alpha \to 0$ and RL to $\alpha \to1$).
We also show that the gradient of the objective function can be considered a mixture of ML- and RL-based objective gradients.
The experimental results of a machine translation task show that minimizing the objective function with $\alpha > 0$ outperforms $\alpha \to 0$, which corresponds to ML-based methods.
\end{abstract}

\section{Introduction}
Neural sequence models have been applied successfully to various types of machine learning tasks, such as neural machine translation~\citep{cho2014learning,sutskever2014sequence,bahdanau2014neural}, caption generation~\citep{xu2015show,chen2015mind}, conversation task~\citep{vinyals2015neural}, and speech recognition~\citep{chorowski2014end,chorowski2015attention,bahdanau2016end}.
As neural sequence models have a wide range of applications, developing more effective and sophisticated learning algorithms can be beneficial.

Popular objective functions for training neural sequence models include the {\it maximum-likelihood} (ML) and {\it reinforcement learning} (RL) objective functions.
However, both have limitations, i.e., training/testing discrepancy and sample inefficiency, respectively.
\citet{bengio2015scheduled} indicated that optimizing the ML objective is not equal to optimizing the evaluation metric (e.g., BLEU~\citep{papineni2002bleu} score in machine translation).
In addition, during training, the ground-truth tokens are used for the prediction of the next token; however, during testing, no ground-truth tokens are available and the tokens that are predicted by the model are used instead.
On the other hand, although the RL-based approach does not suffer from the training/testing discrepancy, it does suffer from sample inefficiency.
Samples generated by the model do not necessarily yield high evaluation scores (i.e., rewards) especially in the early stage of the training.
Consequently, RL-based methods are not self-contained, i.e., they require pre-training via ML-based methods.
As discussed in Section~\ref{sec:objectives}, since these problems depend on their sampling distributions, it is difficult to resolve these problems simultaneously.

We propose $\alpha$-divergence minimization training named \pro{} for a neural sequence model.
We demonstrate that an \pro{} objective function generalizes ML- and RL- based objective functions, i.e., \pro{} can represent both functions as its special cases ($\alpha \to 0$ and $\alpha \to 1$).
We also show that, for $\alpha \in (0, 1)$, the gradient of the \pro{} objective becomes the weighted sum of the gradients of negative log-likelihoods.
Here the weights are obtained by the geometric mean of the sampling distributions of the ML- and RL-based objectives.
We apply gradient descent methods to optimize the proposed objective function, where the gradient of the objective is estimated by means of importance sampling.
Consequently the optimization strategy avoids on-policy RL sampling which suffers from sample inefficiency, and optimizes the objective function closer to the desired RL-based objective.

Experimental results on a machine translation task indicate that
the proposed \pro{} approach outperforms the ML baseline and the reward augmented maximum-likelihood method~(RAML; \citealp{norouzi2016reward}), upon which we build the proposed method.
We compare our results to those reported by \citet{bahdanau2016actor}, who proposed an on-policy RL-based method.
We also confirm that \pro{} can provide comparable BLEU score without pre-training.

The contributions of this paper are summarized as follows.
\begin{itemize}
\item We consider the limitations and advantages of ML, RL, and RAML objective functions with respect to (i)~objective score discrepancy, (ii)~sampling distribution discrepancy and (iii)~sample inefficiency~(Section~\ref{sec:objectives}).
\item We define the \pro{} objective function using $\alpha$-divergence and demonstrate that it can be considered a generalization of ML- and RL-based objective functions~(Section~\ref{sec:proposed-objective}).
\item We demonstrate that the gradient of the \pro{} objective can be obtained by the weighted sum of the gradients of negative log likelihoods and that the weights are a mixture of the sampling distributions of ML- and RL-based objective functions~(Section~\ref{sec:proposed-grad}).
\item We propose an importance sampling-based optimization method, which is very similar to the RAML optimization, for the proposed \pro{} objective function (Section~\ref{sec:optim}). Thus, there is nearly no implementation cost if RAML has already been implemented.
\item The results of machine translation experiments demonstrate that the proposed \pro{} outperforms the ML-baseline and RAML~(Section~\ref{sec:exp}).
\end{itemize}

\section{Comparing objective functions}
\label{sec:objectives}
At least three problems associated with learning neural sequence models exist, and the ML and RL approaches cannot resolve all of these problems simultaneously. In this section, we examine why these problems matter and how  they are addressed by current state-of-the-art methods.

Given a context (or input sequence) $x \in \mathcal{X}$ and a target sequence $y = (y_1, \ldots, y_T) \in \mathcal{Y}$,
an ML approach is typically used to train a neural sequence model.
ML minimizes the negative log-likelihood objective function
\begin{align}
\mathcal{L}(\theta) = - \sum_{x \in \mathcal{X}} \sum_{y \in \mathcal{Y}} q(y | x) \log \ptheta(y|x), \label{eq:L}
\end{align}
where $q(y|x)$ denotes the true sampling distribution.
Here, we assume that $x$ is uniformly sampled from $\mathcal{X}$ and omit the distribution of $x$ from Eq.~\eqref{eq:L} for simplicity.
For example, in machine translation, if a corpus contains only a single target sentence $y^{\ast}$ for each input sentence $x$, then $q(y|x) = \delta(y, y^{\ast} | x)$ and the objective becomes $\mathcal{L}(\theta) = - \sum_{x \in \mathcal{X}} \log \ptheta(y^{\ast}|x)$.

It is known that ML does not optimize the final performance measure.
For example, in the case of machine translation, the famous evaluation measures such as BLEU or edit rate~\citep{snover2006ter} differ from the negative likelihood function.

The optimization of the final performance measure can be formulated as the minimization of the negative total expected rewards expressed as follows:
\begin{align}
\mathcal{L}^{\ast}(\theta) = - \sum_{x \in \mathcal{X}} \sum_{y \in \mathcal{Y}} \ptheta(y | x) r(y, y^{\ast} | x), \label{eq:L*}
\end{align}
where $r(y, y^{\ast}| x)$ is a reward function associated with the sequence prediction $y$, i.e.\ the BLEU score or the edit rate in machine translation.

The above observations raise the following problems.
\begin{enumerate}
\renewcommand{\labelenumi}{(\roman{enumi})}
\item {\bf Objective score discrepancy.} The reward function is not used when training the model; however, it is the performance measure in the testing (evaluation) phase.
\item {\bf Sampling distribution discrepancy.} The model is trained with samples from the true sampling distribution $q(y|x)$; however, it is evaluated using samples generated from the learned distribution $\ptheta(y|x)$.
\end{enumerate}

On-policy RL is an approach to solve the above problems.
The objective function of on-policy RL is $\mathcal{L}^{\ast}$ in Eq.~\eqref{eq:L*}, which is a reward-based objective function; thus, there is no objective score discrepancy, which resolves problem (i).
On-policy sampling from $\ptheta(y|x)$ and taking the expectation with $\ptheta(y|x)$ in Eq.~\eqref{eq:L*} also resolves problem (ii).
\citet{ranzato2015sequence} and \citet{bahdanau2016actor} directly optimized $\mathcal{L}^{\ast}$ using policy gradient methods~\citep{sutton2000policy}.
A sequence prediction task that selects the next token based on an action trajectory $(y_1, \ldots, y_{t-1})$ can be considered an RL problem.
Here the next token selection corresponds to the next action selection in RL.
In addition, the action trajectory and the context $x$ correspond to the current state in RL.
To prevent the policy from becoming overly greedy and deterministic, some studies have used the following entropy-regularized version of the policy gradient objective function~\citep{mnih2016asynchronous}:
\begin{equation}
\mathcal{L}_{(\tau)}^{\ast}(\theta) := \sum_{x \in \mathcal{D}} \Bigl\{ -\tau \mathbb{H}(\ptheta(y|x)) - \sum_{y \in \mathcal{Y}} \ptheta (y|x) r(y, y^{\ast}|x) \Bigr\}. \label{eq:L*tau}
\end{equation}
Note that $\lim_{\tau \to 0}\mathcal{L}_{(\tau)}^{\ast} = \mathcal{L}^{\ast}$ holds.
We can obtain the gradient of the objective as $\nabla_{\theta} \mathcal{L}^{\ast}(\theta) = - \sum_{x \in \mathcal{X}} \sum_{y \in \mathcal{Y}} \ptheta(y | x) \nabla_{\theta} \log \ptheta(y|x) r(y, y^{\ast} | x)$ using the policy gradient theorem~\citep{sutton2000policy}.

On-policy RL can suffer from sample inefficiency; thus, it may not generate samples with high rewards, particularly in the early learning stage.
By definition, on-policy RL generates training samples from its model distribution.
This means that, if model $\ptheta(y|x)$ has low prediction ability, only a few samples will exist with high rewards.
\begin{enumerate}
\renewcommand{\labelenumi}{(\roman{enumi})}
\setcounter{enumi}{2}
\item {\bf Sample inefficiency.} The RL model can draw samples with low rewards, which results in the failure of estimating the objective function around the peak.
\end{enumerate}
Machine translation suffers from this problem because the action (token) space is vast (typically $\mathord{>}10,000$ dimensions) and rewards are sparse, i.e., positive rewards are observed only at the end of a sequence.
Therefore, the RL-based approach usually requires good initialization and thus is {\it not} self-contained.
Previous studies have employed pre-training with ML before performing on-policy RL-based sampling~\citep{ranzato2015sequence,bahdanau2016actor}.

Despite various attempts, a fundamental technical barrier exists.
This barrier prevents solving the three problems using a single method.
The barrier comes from a trade-off between (ii)~sampling distribution discrepancy and (iii)~sample inefficiency because these issues are related to the sampling distribution.

\citet{norouzi2016reward} proposed RAML, which solves problems (i) and (iii) simultaneously.
RAML replaces the sampling distribution of ML, i.e., $q(y | x)$ in Eq.~\eqref{eq:L}, with a reward-based distribution $\qtau(y | x) \propto \exp\left\{ r(y, y^{\ast} | x) / \tau \right\}$.
In other words, RAML incorporates the reward information into the ML objective function.
The RAML objective function is expressed as follows:
\begin{equation}
\mathcal{L}_{(\tau)}(\theta) := - \sum_{x \in \mathcal{X}} \sum_{y \in \mathcal{Y}} \qtau(y|x) \log \ptheta (y | x). \label{eq:Ltau}
\end{equation}
However, problem (ii) remains.
As mentioned above, it is difficult to solve problems (ii) and (iii) simultaneously; thus, our approach is to control the trade-off of the sampling distributions by considering their mixture.

\section{$\alpha$-divergence}
\label{sec:alpha-divergence}

The proposed method utilizes $\alpha$-divergence $D_\mathrm A^{(\alpha)}(p\|q)$, which measures the asymmetric distance between two distributions $p$ and $q$~\citep{shun2012differential}.
A prominent feature of $\alpha$-divergence is that it can behave as $D_\mathrm{KL}(p\|q)$ or $D_\mathrm{KL}(q\|p)$ depending on the value of $\alpha$, i.e., $D_{{\rm A}}^{(1)}(p \| q) := \lim_{\alpha \to 1} D_{{\rm A}}^{(\alpha)}(p \| q) = D_{{\rm KL}}(p \| q)$ and  $D_{{\rm A}}^{(0)}(p \| q) := \lim_{\alpha \to 0} D_{{\rm A}}^{(\alpha)}(p \| q) = D_{{\rm KL}}(q \| p)$.  This fact follows from the definition of $\alpha$-divergence
\begin{align}
D_{{\rm A}}^{(\alpha)}(p \| q)
&:= \frac{1}{\alpha(1-\alpha)} \Bigl\{ 1- \sum_{y \in \mathcal{Y}} p^{\alpha}(y) q^{1-\alpha}(y) \Bigr\}\\
&= - \frac{1}{\alpha} \sum_{y \in \mathcal{Y}} p(y) \glog \biggl(\frac{q(y)}{p(y)}\biggr),
\end{align}
where $\glog(\cdot)$ is the generalized logarithm $\glog(x) := (1-\alpha)^{-1} (x^{1 - \alpha} - 1)$.

\section{\pro}
In this section, we describe the objective function of the proposed method, i.e., \pro, and how it is trained.

\subsection{Objective function}
\label{sec:proposed-objective}
\begin{figure*}[htpb]
  \centering
    \includegraphics[width=15cm]{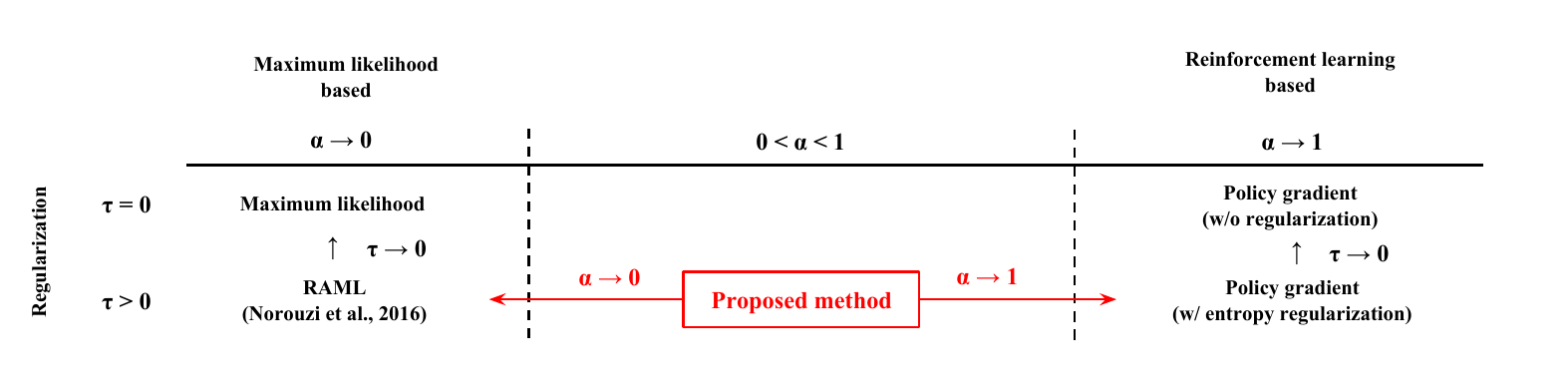}
    \caption{\pro{} objective bridges ML- and RL-based objectives.}
    \label{fig:idea}
\end{figure*}
We define the objective function of \pro{} as the $\alpha$-divergence between $\ptheta$ and $\qtau$:
\begin{align}
\objPro(\theta)
&:= \tau \sum_{x \in \mathcal{X}} D_{{\mathrm A}}^{(\alpha)} (\ptheta \| \qtau)\\
&= - \frac{\tau}{\alpha} \sum_{x \in \mathcal{X}} \sum_{y \in \mathcal{Y}} \ptheta(y|x) \glog\biggl(\frac{\qtau(y|x)}{\ptheta(y|x)} \biggr).  \label{eq:L*at}
\end{align}
This $\alpha$-divergence is equal to (up to constant) $\objEnRL$ in Eq.~\eqref{eq:L*tau} or $\objRAML$ in Eq.~\eqref{eq:Ltau} by employing $\alpha \to 1$ or $\alpha \to 0$ limits, respectively.
Figure~\ref{fig:idea} illustrates how the \pro{} objective bridges the ML- and RL-based objective functions.
\begin{align}
\lim_{\alpha \to 1} \objPro(\theta)
&= \tau \sum_{x \in \mathcal{X}} D_{{\rm KL}}(\ptheta \| \qtau)\\
&= \mathcal{L}_{(\tau)}^{\ast}(\theta) + {\rm constant},\\
\lim_{\alpha \to 0} \objPro(\theta)
&= \tau \sum_{x \in \mathcal{X}} D_{\rm KL}(\qtau \| \ptheta)\\
&= \tau \mathcal{L}_{(\tau)}(\theta) + {\rm constant}.
\end{align}
Although the objectives $\objPro^{\ast}(\theta)$, $\mathcal{L}_{(\tau)}^{\ast}(\theta)$, and $\mathcal{L}_{(\tau)}(\theta)$ have the same minimizer $\ptheta(y|x) = \qtau(y|x)$, empirical solutions often differ.

\subsection{Objective function gradient}
\label{sec:proposed-grad}

The gradient of \eqref{eq:L*at} used for gradient descent optimization in the proposed \pro{} can be obtained by a discussion similar to that of the policy gradient theorem~\citep{sutton2000policy}.
The gradient of the \pro{} objective function is expressed as
\begin{equation}
\nabla_{\theta} \objPro(\theta)
= - \sum_{x \in \mathcal{X}} \sum_{y \in \mathcal{Y}} \disPro(y|x) \nabla_{\theta} \log \ptheta(y|x),
\end{equation}
where
\begin{equation}
\disPro(y|x) = \frac{\tau}{1-\alpha}  \ptheta^{\alpha}(y|x) \qtau^{1-\alpha}(y|x)
\end{equation}
is a weight that mixes sampling distributions $\ptheta$ and $\qtau$.
See Appendix~\ref{sec:appendix-obj} for the derivation of this gradient.
This gradient differs from that of ML or RAML only in the weights.
In addition, it converges to the gradient of RL or RAML (up to constant) by taking $\alpha \to 1$ or $\alpha \to 0$ limits, respectively; i.e.,  $\lim_{\alpha \to 1} \nabla_{\theta} \objPro = \nabla_{\theta} \objEnRL$ and $\lim_{\alpha \to 0}\nabla_{\theta} \objPro = \tau \nabla_{\theta} \objRAML$.

\subsection{Optimization of objective function}
\label{sec:optim}
Our optimization strategy is similar to that of RAML.
First, we estimate the gradient of the objective function by importance sampling and then use this estimate with the gradient method.
We sample target sentence $y$ for each $x$ from a proposal distribution $q_0$ and estimate the gradient by importance sampling as follows:
\begin{multline}
\nabla_{\theta} \objPro(\theta) \\
= - \sum_{x\in\mathcal{X}}\sum_{y \in \mathcal{Y}} q_0(y|x) \biggl( \frac{\disProTilde(y|x)}{q_0(y|x)}\biggr) \nabla \log \ptheta(y|x),
\end{multline}
where $\disProTilde(y|x) = \frac{1}{Z}\ptheta^{\alpha}(y|x)\qtau^{1-\alpha}(y|x)$ is the normalized distribution of $\disPro(y|x)$. This normalization changes the magnitude of the gradient but not the direction.
Here, $q_0$ is typically obtained by applying data augmentation to a corpus and calculating rewards for the generated samples. 

\section{Related works}
From the RL literature, reward-based neural sequence model training can be separated into on-policy and off-policy approaches, which differ in the sampling distributions.
The proposed \pro{} approach can be considered an off-policy approach with importance sampling.

Recently, on-policy RL-based approaches for neural sequence predictions have been proposed.
\citet{ranzato2015sequence} proposed a method that uses the REINFORCE algorithm~\citep{williams1992simple}.
Based on \citet{ranzato2015sequence}, \citet{bahdanau2016actor} proposed a method that estimates a critic network and uses it to reduce the variance of the estimated gradient.
\citet{bengio2015scheduled} proposed a method that replaces some ground-truth tokens in an output sequence with generated tokens.
\citet{yu2016seqgan}, \citet{lamb2016professor}, and \citet{wu2017adversarial} proposed methods based on GAN (generative adversarial net) approaches~\citep{goodfellow2014generative}.
Note that on-policy RL-based approaches can directly optimize the evaluation metric.
\citet{degris2012linear}  proposed off-policy gradient methods using importance sampling, and the proposed \pro{} off-policy approach utilizes importance sampling to reduce the difference between the objective function and the evaluation measure when $\alpha > 0$.

As mentioned previously, the proposed \pro{} can be considered an off-policy RL-based approach in that the sampling distribution differs from the model itself.
Thus, the proposed \pro{} approach has the same advantages as off-policy RL methods compared to on-policy RL methods, i.e., computational efficiency during training and learning stability.
On-policy RL approaches must generate samples during training, and immediately utilize these samples.
This property leads to high computational costs during training and if the model falls into a poor local minimum, it is difficult to recover from this failure.
On the other hand, by exploiting data augmentation, the proposed \pro{} can collect samples before training.
Moreover, because the sampling distribution is a stationary distribution independent of the model, one can expect that the learning process of \pro{} is more stable than that of on-policy RL approaches.
Several other methods that compute rewards before training can be considered off-policy RL-based approaches, e.g., minimum risk training (MRT; \citealp{shen2015minimum}, RANDOMER~\citep{guu2017language}, and Google neural machine translation~(GNMT; \citealp{wu2016google}).

While the proposed approach is a mixture of ML- and RL-based approaches, this attempt is not unique.
The sampling distribution of scheduled sampling~\citep{bengio2015scheduled} is also a mixture of ML- and RL-based sampling distributions.
However, the sampling distributions of scheduled sampling can differ even in the same sentence, whereas ours are sampled from a stationary distribution.
To bridge the ML- and RL-based approaches, \citet{guu2017language} considered the weights of the gradients of the ML- and RL-based approaches by directly comparing both gradients.
In contrast, the weights of the proposed \pro{} approach are obtained as the results of defining the $\alpha$-divergence objective function.
GNMT~\citep{wu2016google} considered a mixture of ML- and RL-based objective functions by the weighted arithmetic sum of $\objML$ and $\objRL$.
Comparing this weighted mean objective function and \pro{}'s objective function could be an interesting research direction in future.


\section{Numerical experiments}
\label{sec:exp}
\begin{table}[tbp]
  \caption{{\bf IWSLT'14 German--English machine translation performance:} The best BLEU scores for the development set and corresponding BLEU scores for the test set are shown. Each search algorithm is greedy or beam search (BS) (beam size 10). AC+ML and RF-C+ML denote actor-critic + maximum likelihood and REINFORCE-critic + maximum likelihood, respectively. Both combine on-policy RL- and ML-based objective functions~\citep{bahdanau2016actor}.}
  \label{table:peformance-iwslt}
\centering
\small
\begin{tabular}{lcccc}
\toprule
BLEU &  Dev (greedy) & Test (greedy) & Test (BS) \\ \midrule
\multicolumn{4}{c}{Results from \citet{bahdanau2016actor}} \\ \midrule
ML & n/a & 25.82 & 27.56\\
AC + LL & n/a & 27.49 & {\bf 28.53}\\
RF-C+LL & n/a & {\bf 27.7} & 28.3\\ \midrule
\multicolumn{4}{c}{Our results} \\ \midrule
ML & 29.83 & 27.96 & 28.26\\
RAML & 29.65 & 27.50 & 28.35 \\
Ours ($\alpha=0.3$)& 29.90 & 27.73 & 28.29 \\
Ours ($\alpha=0.5$)& {\bf 29.91} & {\bf 28.02} & {\bf 28.49} \\
Ours ($\alpha=0.7$)& 29.72 & 27.81 & 28.25 \\
\bottomrule
\end{tabular}
\end{table}
We evaluated the effectiveness of \pro{} experimentally using a neural machine translation task.
We compared the BLEU scores of the ML baseline, RAML, and the proposed \pro{} on the IWSLT'14 German--English corpus~\citep{cettolo2014report}.
We trained the same attention-based encoder-decoder model~\citep{bahdanau2014neural,luong2015effective} for each method.
When sampling from $\qtau$, we employed a data augmentation procedure similar to that of \citet{norouzi2016reward}.
As a result, we found that \pro{} with $\alpha > 0$ without pre-training outperformed the ML baseline and RAML.

To compare our experimental results to on-policy RL methods~\citep{ranzato2015sequence,bahdanau2016actor}, we used the IWSLT'14 German--English corpus~\citep{cettolo2014report} and an attention-based encoder-decoder model~\citep{bahdanau2014neural,luong2015effective}.
The training data comprised 150K German--English sentence pairs and approximately 7K development/test sentence pairs.
We followed the model parameters used by \citet{bahdanau2016actor} and used the same encoder-decoder model architecture for all methods.
Details about the models and parameters are discussed at the end of this section.

We obtained augmented data in the same manner as the RAML framework~\citep{norouzi2016reward}.
For each target sentence, some tokens were replaced by other tokens in the vocabulary and we used the negative Hamming distance as reward.
We assumed that Hamming distance $e$ for each sentence is less than $[m \times 0.2]$, where $m$ is the length of the sentence and $[a]$ denotes the maximum integer which is less than or equal to $a \in \mathbb{R}$.
Moreover, the Hamming distance for a sample is uniformly selected from $0$ to $[m \times 0.2]$.
One can also use BLEU or another machine translation metric for this reward.
However, we assumed the different proposal distribution $q_0$ from that of RAML.
We assumed the simplified proposal distribution $q_0$, which is a discrete uniform distribution over $[0, m \times 0.2]$.
This results in that hyperparameter $\tau$ used in this experiment is larger than that of RAML; thus, $\tau$ was set to $3.0$ for RAML and the proposed \pro{}.

The experimental results obtained with the IWSLT'14 corpus are shown in Table~\ref{table:peformance-iwslt}.
We calculated the BLEU scores with {\it multi-bleu.perl}\footnote{\url{https://github.com/moses-smt/mosesdecoder/blob/master/scripts/generic/multi-bleu.perl}} script for both the development and test sets.
As shown in Table~\ref{table:peformance-iwslt}, the best BLEU score on the development set with greedy search prediction and the corresponding BLEU scores on the test set with greedy and beam search prediction are shown for each method.
Here, the beam width was set to 10~\citep{ranzato2015sequence,bahdanau2016actor}.
We found that the proposed \pro{} with $\alpha = 0.5$ outperformed the ML-baseline and RAML.
This implies that an objective function better than RAML for neural sequence model training exists in $\alpha > 0$.
Note that the proposed \pro{} does not utilize pre-training; on the other hand, an on-policy RL approach~\citep{bahdanau2016actor} needs good initialization by pre-training using ML.
Although the ML baseline performances differ between our results and those of \citet{bahdanau2016actor},
 we emphasize that the proposed \pro{} performance with $\alpha = 0.5$ without pre-training is comparable with the on-policy RL-based methods.

\paragraph{Model details}
 The model architecture and parameters follow that of \citet{ranzato2015sequence} and \citet{bahdanau2016actor}.
 Here, the encoder was a bidirectional LSTM with 256 units.
 The decoder was also an LSTM with the same number of units.
 The vocabulary sizes for the source/target were 32\,009 and 22\,822, respectively.
 We utilized stochastic gradient descent with a decaying learning rate.
 The learning rate decays from $1.0$ to $0.05$ with dev-decay~\citep{wilson2017marginal}, i.e., after training each epoch, we monitored the greedy BLEU score for the development set and reduced the learning rate by multiplying it with $\delta = 0.5$ only when the greedy BLEU score for the development set did not update the best BLEU score.
 The minibatch size was 128.
 In addition, if an unknown token, i.e.\ a special token representing a word not in the vocabulary, was generated in the predicted sentence, it was replaced by the token with the highest attention in the source sentence~\citep{jean2014using}.
 We implemented our models using a fork from the PyTorch\footnote{\url{http://pytorch.org}} version of the OpenNMT toolkit~\citep{klein2017opennmt}.
 Our reference implementation is available: \url{https://github.com/sotetsuk/alpha-dimt-icmlws}.

\section{Conclusion}
In this study, we have proposed a new objective function as $\alpha$-divergence minimization for neural sequence model training that unifies ML- and RL-based objective functions.
In addition, we proved that the gradient of the objective function is the weighted sum of the gradients of negative log-likelihoods, and that the weights are represented as a mixture of the sampling distributions of the ML- and RL-based objective functions.
We demonstrated that the proposed approach outperforms the ML baseline and RAML in the IWSLT'14 machine translation task.

By extending the existing objectives via $\alpha$-divergence, we gained an additional freedom to control the trade-off between the training/testing discrepancy and sample inefficiency.
We consider that the sample inefficiency disappears as the learning of the neural sequence model progresses if $\alpha$ is increased as the learning progresses.
We will investigate methods to tune the control parameter $\alpha$ in accordance with the learning process in future.

\section*{Acknowledgments}

AK was supported in part by NEDO, Japan, and JSPS KAKENHI 26730130.

{\small
\bibliography{reference}
\bibliographystyle{icml2017}
}

\appendix

\section{Gradient of \pro{} objective}
\label{sec:appendix-obj}
The gradient of \pro{} can be obtained as follows:
\begin{align}
&\nabla_{\theta} \mathcal{L}_{(\alpha, \tau)}^{\ast}(\theta) \notag\\
&= \nabla_{\theta} \Bigl\{ - \sum_{x \in \mathcal{X}} \frac{\tau}{\alpha(1-\alpha)} \Bigl\{ 1- \sum_{y \in \mathcal{Y}} \ptheta^{\alpha}(y|x) \qtau^{1-\alpha}(y|x) \Bigr\} \Bigr\} \\
&= - \frac{\tau}{\alpha(1-\alpha)} \sum_{x \in \mathcal{X}} \sum_{y \in \mathcal{Y}} \nabla_{\theta} \ptheta^{\alpha}(y|x) \qtau^{1-\alpha}(y|x)\\
&= - \frac{\tau}{1-\alpha} \sum_{x \in \mathcal{X}} \sum_{y \in \mathcal{Y}} \ptheta^{\alpha}(y|x) \qtau^{1-\alpha}(y|x) \nabla_{\theta} \log \ptheta(y|x) \label{exp:log_trick}\\
&= - \sum_{x \in \mathcal{X}} \sum_{y \in \mathcal{Y}} \disPro(y|x) \nabla_{\theta} \log \ptheta(y|x),
\end{align}
where
\begin{equation}
\disPro(y|x) = \frac{\tau}{1-\alpha}  \ptheta^{\alpha}(y|x) \qtau^{1-\alpha}(y|x).
\end{equation}
In Eq.~\eqref{exp:log_trick}, we used the so-called {\it log-trick}: $\nabla_{\theta} \ptheta(y|x) = \ptheta(y|x) \nabla_{\theta} \log \ptheta(y|x)$.
\end{document}